\documentclass[runningheads]{llncs}
\usepackage[T1]{fontenc}
\usepackage{amsmath}
\usepackage{graphicx}
\usepackage{subcaption}

\usepackage{booktabs}
\usepackage{array}
\usepackage{multirow}
\usepackage{comment}

\begin{document}
\title{FedDPG: An Adaptive Yet Efficient Prompt-tuning Approach in Federated Learning Settings}
\titlerunning{FedDPG}

\author{Ali Shakeri\inst{1} \and
Wei Emma Zhang\inst{1}\orcidID{000-0002-0406-5974} \and \\
Amin Beheshti \inst{2}\orcidID{0000-0002-5988-5494} \and
Weitong Chen \inst{1}\orcidID{0000-0003-1001-7925} \and \\
Jian Yang  \inst{2}\orcidID{0000-0002-4408-1952} \and
Lishan Yang \inst{1}
}
\authorrunning{Zhang et al.}
%
\institute{The University of Adelaide, Adelaide, Australia\and
Macquarie University, Sydney, Australia \\
\email{\{ali.shakeri,wei.e.zhang\}@adelaide.edu.au}\\
}

\maketitle              
\begin{abstract}
Pre-trained Language Models (PLMs) have demonstrated impressive performance in various NLP tasks. However, traditional fine-tuning methods for leveraging PLMs for downstream tasks entail significant computational overhead. Prompt-tuning has emerged as an efficient alternative that involves prepending a limited number of parameters to the input sequence and only updating them while the PLM's parameters are frozen. However, this technique's prompts remain fixed for all inputs, reducing the model's flexibility. The Federated Learning (FL) technique has gained attention in recent years to address the growing concerns around data privacy. However, challenges such as communication and computation limitations of clients still need to be addressed. To mitigate these challenges, this paper introduces the \textbf{Fed}erated \textbf{D}ynamic \textbf{P}rompt \textbf{G}enerator (FedDPG), which incorporates a dynamic prompt generator network to generate context-aware prompts based on the given input, ensuring flexibility and adaptability while prioritising data privacy in federated learning settings. Our experiments on three NLP benchmark datasets showcase that FedDPG outperforms the state-of-the-art parameter-efficient fine-tuning methods in terms of global model performance\footnote{Compared with five models on three datasets, with only one configuration has marginal lower performance}, and has significantly reduced the calculation time and the number of parameters to be sent through the FL network.

\end{abstract}

\section{Introduction}
Large Language Models (LLMs) have shown great potential in various Natural Language Processing (NLP) tasks; however, training these models requires a large amount of data, which has heightened concerns about data privacy. Additionally, training LLMs require powerful computational resources, making it more difficult for small to medium organisations to invest in training one for their business. The federated learning (FL) technique, proposed by McMahan et al. \cite{mcmahan-fedavg}, offers a solution by enabling AI model training, including training LLMs, without sharing private data. FL facilitates collaboration among multiple organisations to train AI models while ensuring their datasets remain localised and protected, even in scenarios of limited data availability. Organisations across various industries can harness FL to train AI models while prioritising data privacy collaboratively. For example, in the telecommunications industry, companies can jointly develop advanced scam detection models powered by LLMs. This approach ensures robust protection for users while maintaining the confidentiality of their data \cite{fl-for-dataproduct}.
Extensive research works have explored FL from various perspectives and challenges, such as FL architectures and concepts \cite{fl_concepts}, data privacy preservation challenges \cite{ppfl}, communication costs \cite{mobile_comm}, the impact of data distribution (IID vs. non-IID) \cite{non-iid-survey}, and Personalized Federated Learning (PFL), which focuses on customising the global model for each client \cite{pfl-survey}. 
Additionally, Large Language Models (LLMs) have emerged as a promising research direction within Federated Learning (FL) systems \cite{prefix-tuning,fedprompt,fedpeptao}, attracting increasing attention in recent years.

Due to the vast number of parameters in LLMs, training them from scratch incurs substantial computational costs. A common way is to fine-tune the Pre-trained Language Models (PLMs) for a downstream task, which could update a part of the PLM parameters and thus is more efficient than training them from scratch \cite{bart,transfer-learning-fine-tuning}. However, fine-tuning still requires accessing the PLM's parameters, which can exceed hundreds of billions \cite{llama3.1}. Parameter-efficient fine-tuning (PEFT) \cite{peft-1,peft-2,peft-3} has recently emerged as a promising approach for customising pre-trained language models (PLMs). This approach involves integrating a small set of trainable parameters into the PLM, either by incorporating them as part of the PLM's input (soft prompts) \cite{prompting-3} or by embedding small neural network modules within the PLM's architecture \cite{adapter-tuning} while keeping the original model parameters frozen. Compared to traditional fine-tuning techniques, PEFT methods can significantly reduce the computational cost, decreasing the number of trainable parameters by up to 90\% \cite{adapter-tuning}.

Integrating PEFT methods into FL systems has garnered significant attention recently as researchers explore practical solutions for efficiently customising PLMs for various downstream tasks \cite{peft-fl-survey}. This approach is particularly promising because it freezes most of the PLM's parameters, requiring only the transfer of newly added parameters. As a result, it reduces the computational demands on clients and addresses key communication and computation challenges associated with FL. For example, FedPrompt \cite{fedprompt} adapts the soft prompt-tuning method for FL, enabling the exchange of trained prompts between the server and clients. Similarly, FedPepTAO \cite{fedpeptao} introduces a scoring mechanism to selectively transfer trained prompts, optimising the communication process within FL systems.

Building on current state-of-the-art PEFT methods, particularly prompt-tuning methods, we propose the \textbf{Fed}erated \textbf{D}ynamic \textbf{P}rompt \textbf{G}enerator (FedDPG) to leverage the capabilities of PLMs in FL settings. We design FedDPG to be more flexible and dynamic than existing PEFT methods while maintaining high performance. Unlike existing prompt-tuning methods, which train a fixed set of prompt vectors and apply the same vectors to all inputs during inference, FedDPG incorporates an auxiliary network to generate unique prompt vectors for each input dynamically. These input-specific vectors encapsulate general information about the input, enhancing the performance of PLMs in text classification tasks. Given the paramount importance of the "Right to Be Forgotten" (RTBF) in machine learning and growing privacy concerns, we also explore the emerging domain of Federated Machine Unlearning (FMU). Through this exploration, we propose FedDPGu, which introduces a novel approach to federated unlearning within PEFT methods.





\section{Related Works}



\subsection{Parameter-efficient Fine-tuning of Large Language Models}
Rebuffi et al. \cite{adapter-tuning} proposed adapter-tuning, which inserts small trainable neural network modules, known as adapters, into the model architecture. Similarly, Zhang et al. \cite{side-tuning} introduced Side-tuning, which incorporates a separate trainable side network combined with the frozen base model through summation. In a related approach, Li et al. \cite{prefix-tuning} proposed Prefix-tuning that adds an auxiliary network, specifically a multi-layer perceptron (MLP) like Side-tuning. However, this network is only used to train continuous vectors. 
LoRA \cite{lora} is another PEFT method for adapting PLMs by injecting trainable low-rank matrices into the transformer model layers while its main parameters are frozen, which significantly reduces the number of trainable parameters by 10,000 times. Discrete prompting \cite{prompting-1} is an effective method for guiding PLMs to produce desired outputs by providing task-specific instructions and contextual examples in plain text, typically in a manual format. This approach is particularly efficient, as it eliminates the need to modify the model's parameters while still achieving competent performance in certain scenarios. However, its performance often falls short compared to fine-tuning and other PEFT techniques \cite{prompting-3}. Continuous prompt-tuning \cite{soft-prompting,p-tuning} has emerged as an effective alternative to discrete prompting. This method involves prepending continuous, trainable parameters (i.e., soft prompts) to the input sequence while keeping the PLM’s parameters fixed and updating only the soft prompts during training. Building on this concept, Liu et al. \cite{p-tuning-v2} introduced P-tuning v2, which employs deep prompt tuning for PLMs by extending prompt-tuning by integrating distinct soft prompts into every layer of the language model rather than confining them to the input layer.

\subsection{Federated Learning for Parameter-efficient Fine-tuning Methods}

The increasing demand for large language models (LLMs) and rising concerns over data privacy have brought attention to developing effective federated learning (FL) systems for tailoring LLMs to downstream tasks. FedPrompt \cite{fedprompt} extends the original prompt-tuning method to FL systems, reducing the computational and communication costs of fine-tuning PLMs by updating only the prompts on clients and aggregating them on the server. Zhao et al. \cite{zhao2024breaking} introduced a multilingual federated prompt-tuning framework to train robust multilingual models for low-resource languages while preserving data privacy efficiently. Che et al. \cite{fedpeptao} proposed FedPepTAO, which adapts P-tuning v2 for FL and incorporates a scoring mechanism on clients that evaluates the correlation between layer-specific prompts and their outputs, enabling the aggregation of only the most critical prompts while leaving other locally trained prompts untouched on each client. Sun et al. \cite{fedbpt} developed FedBPT, a method designed to reduce the computational burden on resource-constrained clients by employing the Covariance Matrix Adaptation Evolution Strategy (CMA-ES) technique. This gradient-free optimization approach allows clients to perform only inference with the PLM and forward propagation during local training, optimizing prompt distributions based on private data.


\section{Methodology}

In this section, we first revisit the problem formulation of federated learning, followed by the introduction of our proposed method FedDPG, which employs prompt-tuning for federated learning in a dynamic way. 

\subsection{Problem Settings}

\subsubsection{NLP Tasks.}
We focus on text classification tasks for our experiments in which input $x$ is a sequence of token embeddings $[e_1, e_2, \dots, e_n]$ (e.g., a movie review) and output $y$ is a label indicating the class of the $x$.

\subsubsection{Dynamic prompt generation.}
To the best of our knowledge, state-of-the-art PEFT methods primarily train a set of parameters, such as soft prompts or prefixes, which are subsequently used as static vectors tailored to specific tasks. In contrast, we propose a dynamic and adaptable system that adjusts prompts based on the given input sequence. This approach enhances LLM performance during inference on unseen data by automatically extracting contextual information from the input and integrating it with the model as soft prompts. Our system employs a neural network preceding the LLM to generate prompts dynamically for each input. We denote the set of dynamic prompts as $P$, the number of prompt vectors (prompt length) as $|P|$, and treat $|P|$ as a hyper-parameter that can be tuned for optimal performance.

\subsubsection{Federated learning system.}
Federated learning systems can be broadly categorized into centralized (CFL) and decentralized (DFL) frameworks \cite{cfl-vs-dfl}. The key difference lies in how global model aggregation is performed among participating clients. In CFL, a central server orchestrates communication and aggregation processes, serving as a hub for coordinating updates from clients. In contrast, DFL eliminates the need for a central server, relying instead on alternative mechanisms for aggregation and communication. For instance, one client can assume a coordination role, facilitating aggregation tasks among other participants. We follow the CFL architecture  
that utilises the Equation (\ref{eq:aggregation}) for performing model aggregation (FedAvg \cite{mcmahan-fedavg}):

\begin{equation}
\label{eq:aggregation}
\mathcal{M}_{global}^{t+1} = \frac{1}{C} \sum_{c=1}^{C} \mathcal{M}_c^t
\end{equation}
where $C$ is the total number of selected clients at round $t$. This method ensures that all clients contribute equally to the global model update, regardless of factors such as dataset size or task performance.


\begin{figure}[t]
    \centering
    \includegraphics[width=0.85\linewidth]{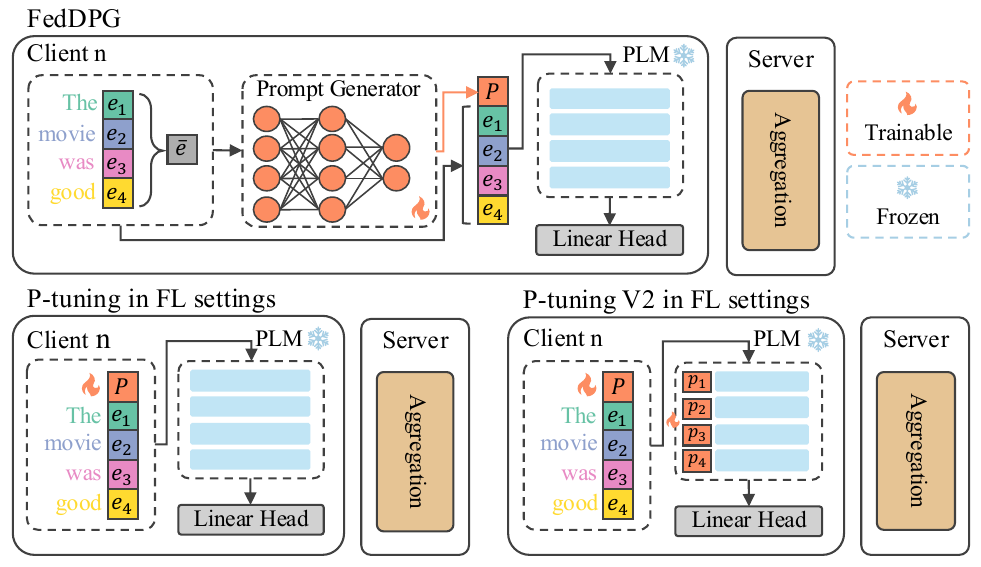}
    \caption{Comparison of the FedDPG architecture with P-tuning \cite{p-tuning} and P-tuning V2 \cite{p-tuning-v2}. FedDPG computes the input sequence's average embedding to generate a context-aware soft prompt ($P$), which is prepended to the input and processed by a frozen PLM. In P-tuning, randomly initialized trainable prompts ($P$) are prepended to the input, while P-tuning V2 extends this with layer-specific prompts ($[p_1, p_2, \dots, p_l]$). 
    All methods aggregate only trainable parameters during federated training rounds, keeping PLM parameters local.}
    \label{fig:FedDPG-overview}
\end{figure}

\subsection{Federated Dynamic Prompt Generator (FedDPG)}

Figure \ref{fig:FedDPG-overview} illustrates the architecture of FedDPG and compares it with P-tuning \cite{p-tuning} and P-tuning V2 \cite{p-tuning-v2}. The primary distinction of FedDPG is its prompt generator network, $\mathcal{G}$, which generates context-aware prompts to improve the PLM's classification accuracy. While $\mathcal{G}$ can use sequential architectures such as LSTM, GRU, or transformers, these models are parameter-intensive and less efficient to train compared to a multi-layer perceptron (MLP). To maintain FedDPG's parameter efficiency and reduce its memory and computational footprint, we adopt an MLP architecture for $\mathcal{G}$.

\subsubsection{Architecture of $\mathcal{G}$.}

To enable the MLP to generate prompts tailored to the input sequence, first the average embedding of the input sequence is calculated as $\Bar{e} = \frac{1}{n} \sum_{i=1}^{n} e_i$, where $n$ is the number of tokens in the sequence. The $\Bar{e}$ is then fed into $\mathcal{G}$ to produce dynamic prompts, $P = \mathcal{G}(\Bar{e})$. $\mathcal{G}$ is a two-layer MLP network with an input dimension equal to the dimension of the word embeddings, $d_{in} = d_e$. Its output dimension is the product of the embedding dimension and the number of prompts, $d_{out} = d_e \times |P|$. To properly form the set of prompts, we reshape $P$ as follows:  

\begin{equation}
\label{eq:reshape}
P_{(|P|, d_e)} = \text{reshape}(P_{(1, d_{out})})
\end{equation}  

The reshaped $P$ is then prepended to $x$ as follows:
\begin{equation}
\label{eq:z}
z=[P_{(|P|, d_e)};x_{(n, d_e)}]
\end{equation}

Finally, the combined embeddings, $z$, are passed to the PLM for classification, yielding the output $\hat{y} = PLM(z)$.

\subsubsection{Local Training and Aggregation of $\mathcal{G}$.}  
Since the PLM's parameters remain frozen during training, each client trains only $\mathcal{G}$ by minimising the following loss function to optimise $\mathcal{G}$:  
\begin{equation}  
\label{eq:loss}  
\mathcal{L}_c = \sum_{i=1}^{n_c} \ell(\theta_P, y_c, \hat{y}_c)
\end{equation}  
where $\mathcal{L}_c$ represents the total local loss for client $c$, $n_c$ is the total number of local data samples, $y_c$ and $\hat{y}_c$ are the ground truth label and the model's prediction respectively, $\theta_P$ is the local trainable prompts, and \(\ell(\cdot, \cdot)\) is the loss function (e.g., cross-entropy). 
This design allows us to safely exclude the PLM's parameters from consideration and perform aggregation solely for $\mathcal{G}$ using Equation (\ref{eq:aggregation}). By doing so, we significantly reduce communication and computation costs while maintaining the high performance of FedDPG.

\subsection{FedDPG for Unlearning (FedDPGu)}
\label{sec:sub_unlearning}



Consider a client $ c $, which seeks to "unlearn" data point $ (d_c^u, y_c^u) $, i.e., a data point that was previously used in training but is now required to be removed from the model’s learned representation. Our methodology for unlearning follows a modified version of the unlearning approach to Learning with Random Labels. 
Formally, suppose the global prompt generator $\mathcal{G}$ at iteration $t$ is denoted as $\mathcal{G}_{global}^t$. To initiate unlearning locally, the client first performs a random relabeling of $ y_c^u $, assigning a randomly chosen label $ \hat{y}_c^u $ from the label space $\mathcal{Y}$, such that $\hat{y}_c^u \neq y_c^u$. 
This perturbed data tuple $ (d_c^u, \hat{y}_c^u) $ is then incorporated into the client's local prompt tuning mechanism, which updates the local prompt generator $\mathcal{G}_c$. The client refines its model to minimise the following localised loss with respect to both the relabeled data and some randomly selected correct samples from its private data: 

\begin{equation}
\label{eq:unlearn_optimisation}
\mathcal{L}_{cu} = \sum_{i=1}^{n_c} \left( \ell(\theta_P, y_c^i, \hat{y}_c^i) + \lambda \ell(\theta_P, y_c^u, \hat{y}_c^u) \right)
\end{equation}
where $\mathcal{L}_{cu}$ represents the FedDPGu's total local loss for client $c$, $n_c$ is the total number of local data samples, $\ell(\theta_P, y_c, \hat{y}_c$ denotes the loss for the original data sample with the correct labelling, $\ell(\theta_P, y_c^u, \hat{y}_c)$ represents the loss for a perturbed data sample, and \( \lambda \) is a regularisation coefficient that balances the contributions from the correct and relabeled losses. Note that $y_c$ and $y_c^u$ represent the ground truth and relabeled labels respectively and $\hat{y_c}$ is the prediction.

\subsubsection{Model Aggregation.}
Once unlearning has been executed locally, the client transmits the updated $\mathcal{G}_c^u$. Unlike traditional federated aggregation protocols such as Equation (\ref{eq:aggregation}), in the unlearning process, the server replaces $\mathcal{G}_{global}$ with the unlearned model provided by the client.
Thus, at the next iteration, the global $\mathcal{G}_{global}(\theta^{t+1})$ directly reflects the updates from the client that performed the unlearning, $\mathcal{G}_{global}^{t+1} \leftarrow \mathcal{G}_c^u$, where $\mathcal{G}_c^u$ represents the client \(c\)'s updated prompt generator and $\mathcal{G}_{global}^{t+1}$ is the new global prompt generator on the server. Because this is a direct replacement rather than aggregation, no other clients need to participate, and the impact of unlearning is confined to the specific data point(s) forgotten by the requesting client.





\section{Experiment}

In this section, we present our experimental setup and evaluate the performance of our proposed model, FedDPG. 

\subsection{Setup}

\textbf{Hardware and Tools.} We simulated the federated learning environment on a single workstation equipped with an Nvidia GeForce RTX 3090 GPU. We set the number of clients to 100 and trained the model with varying client selection ratios and prompt lengths. We implemented the model using PyTorch \cite{pytorch} and utilised Hugging Face \cite{huggingface} to integrate PLMs into FedDPG.
\textbf{Datasets.} In this experiment, we chose three widely used datasets for sentiment analysis and text classification tasks; 1) The Stanford Sentiment Treebank (SST-2) \cite{sst2}, which consists of sentences from movie reviews along with human annotations of their sentiment. 2) AG News \cite{ag_yelp}, a collection of over 1 million news articles gathered from more than 2,000 news sources by ComeToMyHead over a period of one year. 3) Yelp Polarity \cite{ag_yelp}, a dataset from the Yelp Dataset Challenge in 2015. This dataset contains over 1.5 million reviews from the Yelp website.
\textbf{Language Models.} We integrated RoBERTa Base with 125 million parameters \cite{roberta}, into FedDPG. RoBERTa (Robustly Optimized BERT Approach) is a transformer-based model widely used for various NLU tasks. Despite the availability of RoBERTa Large, which boasts 355 million parameters and offers enhanced performance, we opted for RoBERTa Base with 125 million parameters to optimise our computational efficiency and reduce training duration.
\textbf{Baseline methods.} We compared four existing PEFT techniques in FL settings as our baselines: FedPrompt \cite{fedprompt}, P-tuning V2 \cite{p-tuning-v2}, FedPepTAO \cite{fedpeptao}, and FedBPT \cite{fedbpt}. Additionally, we included the classic FL method, FedAvg \cite{mcmahan-fedavg}, as part of our baseline comparisons.

\subsection{Results}

\subsubsection{Global performance.}
Table \ref{tab:global-performance} compares the global performance of FedDPG with the baseline methods, demonstrating FedDPG's superiority on the AG News and Yelp Polarity datasets. While the accuracy of Fed P-tuning and FedPrompt slightly exceeds FedDPG on the SST-2 dataset by 0.46\% and 0.11\%, respectively, these methods utilise RoBERTa Large as their backbone PLM, whereas FedDPG uses RoBERTa Base, which has 230 million fewer parameters. Moreover, despite their marginal advantage on SST-2, these methods perform worse than FedDPG on the AG News and Yelp Polarity datasets, highlighting FedDPG's overall robustness and effectiveness across diverse tasks.


\begin{table} [t]
    \caption{Comparison of global model accuracy (\%) of FedDPG and baseline methods across various datasets in FL settings}\label{tab:global-performance}
    \centering
    \begin{tabular}{|l|c|c|c|c|c|}
        \hline
        Method & PLM & Params. & SST-2 & AG News & Yelp Polarity  \\ \hline
        FedPrompt & RoBERTa Large & 51k & 90.25 & 87.72 & 91.44 \\ 
        Fed P-tuning V2 & RoBERTa Large & 15M & \textbf{90.6} & 88.17 & 93.61 \\ 
        FedAvg & RoBERTa Large & 355M & 84.7 & 77.43 & 88.25 \\ 
        FedPepTAO & RoBERTa Base & 368K & 86.46 & 87.75 & 93.04 \\ 
        FedBPT & RoBERTa Large & 500 & 87.16 & 82.36 & 91.12 \\ \hline 
        \textbf{FedDPG (Ours)} & \textbf{RoBERTa Base} & 185K & 90.14 & \textbf{92.33} & \textbf{94.95} \\ \hline
    \end{tabular}
\end{table}


\subsubsection{Evaluating the Effectiveness of $\mathcal{G}$}
\begin{figure}
    \centering
    \includegraphics[width=1\linewidth]{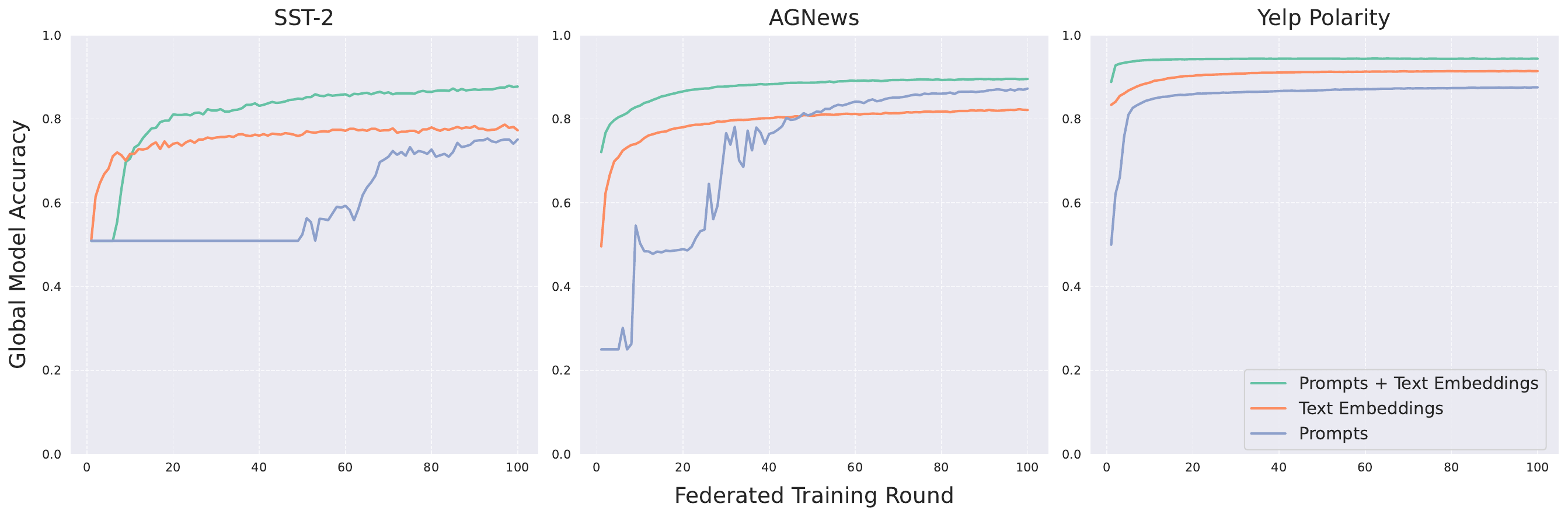}
    \caption{Global model performance across 100 federated rounds using three input configurations: (1) $[P; x]$ (text embeddings with generated prompts), (2) $x$ (text embeddings only), and (3) $P$ (generated prompts only).}
    \label{fig:performance_comparison_mlp}
\end{figure}

To evaluate the effectiveness of $\mathcal{G}$ in dynamically generating prompts $P$, we trained FedDPG using three different input configurations for its PLM: (1) $z = [P; x]$, as described in the methodology section; (2) bypassing $\mathcal{G}$ entirely and feeding the PLM directly with $x$; and (3) using only the generated prompts $P$ as input to the PLM. Figure \ref{fig:performance_comparison_mlp} highlights the contribution of $\mathcal{G}$, demonstrating its role in enhancing the accuracy of FedDPG by smoothing and accelerating its performance.

\subsubsection{Influence of $\mathcal{G}$ Parameter Size.}
\begin{figure}
    \centering
    \includegraphics[width=0.6\linewidth]{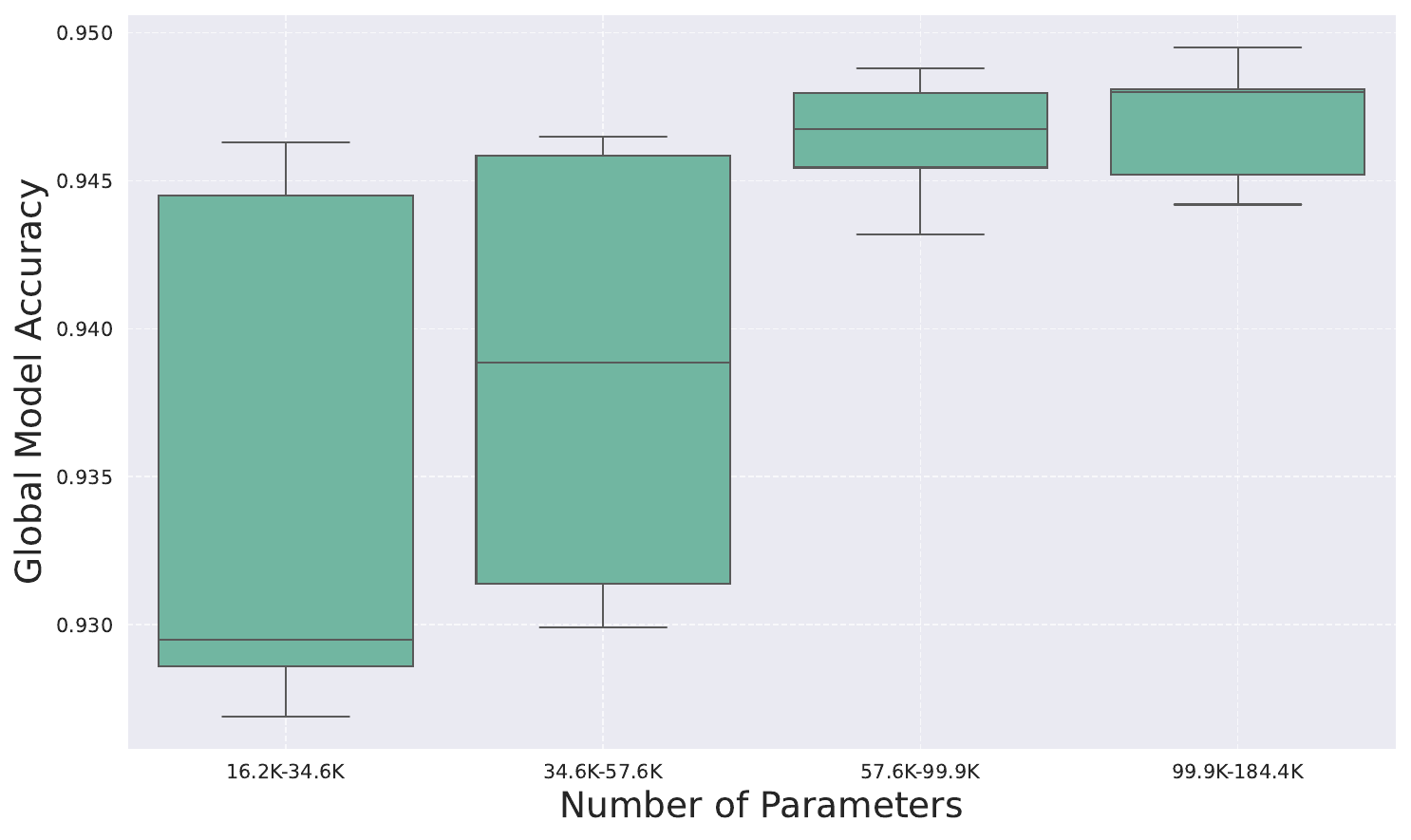}
    \caption{Influence of $\mathcal{G}$ parameter size over global model's performance on Yelp Polarity dataset}
    \label{fig:performance_vs_params}
\end{figure}

To evaluate the impact of the number of parameters in $\mathcal{G}$ on FedDPG's performance, we set its hidden dimensions to 5, 10, and 20. Figure \ref{fig:performance_vs_params} demonstrates the expected trend: larger models tend to perform better. Additionally, the figure highlights that the size of $\mathcal{G}$ has a greater influence on performance compared to other hyperparameters, such as prompt length and client selection ratio. This is evident from the reduced variability (smaller box sizes) in the boxplot as the number of parameters increases.

\vspace{-10mm}

\subsubsection{Impact of Prompt Length and Client Selection Ratio.}
We conducted experiments with different combinations of client selection ratios (5\%, 10\%, and 20\%) and prompt lengths (1, 5, and 10) with $\mathcal{G}$ hidden dimension set to 10. The results demonstrated that increasing both the client selection ratio and prompt length generally improved model performance across all three datasets. The best performance was achieved with 20\% client selection, where the model reached accuracies of 89.33\% on SST-2, 91.93\% on AG News, and 94.88\% on Yelp Polarity using 5 or 10 prompt vectors. Notably, increasing prompt length from 1 to 5 vectors showed substantial improvements (e.g., from 80.85\% to 87.39\% on SST-2 with 5\% client selection), while the gain from 5 to 10 vectors was relatively marginal.
\vspace{-2mm}

\subsubsection{FedDPGu.}

\begin{figure}[!ht]
    \centering
    \begin{subfigure}[t]{0.45\linewidth}
        \centering
        \includegraphics[width=\linewidth]{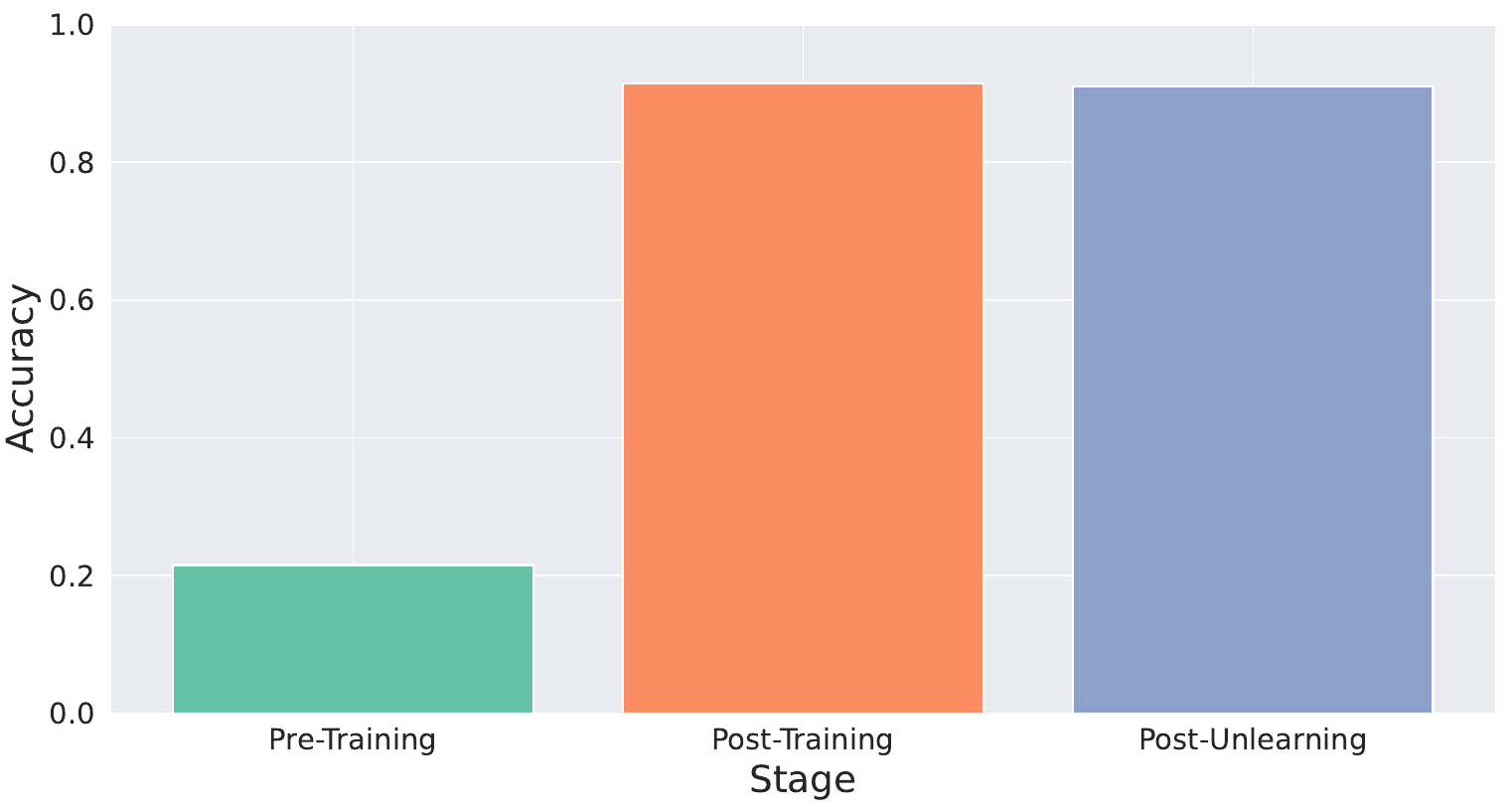}
        \caption{The performance of the global model at the start of the training, before, and after unlearning process (From left to right)}
        \label{fig:fu_global}
    \end{subfigure}
    \hfill
    \begin{subfigure}[t]{0.45\linewidth}
        \centering
        \includegraphics[width=\linewidth]{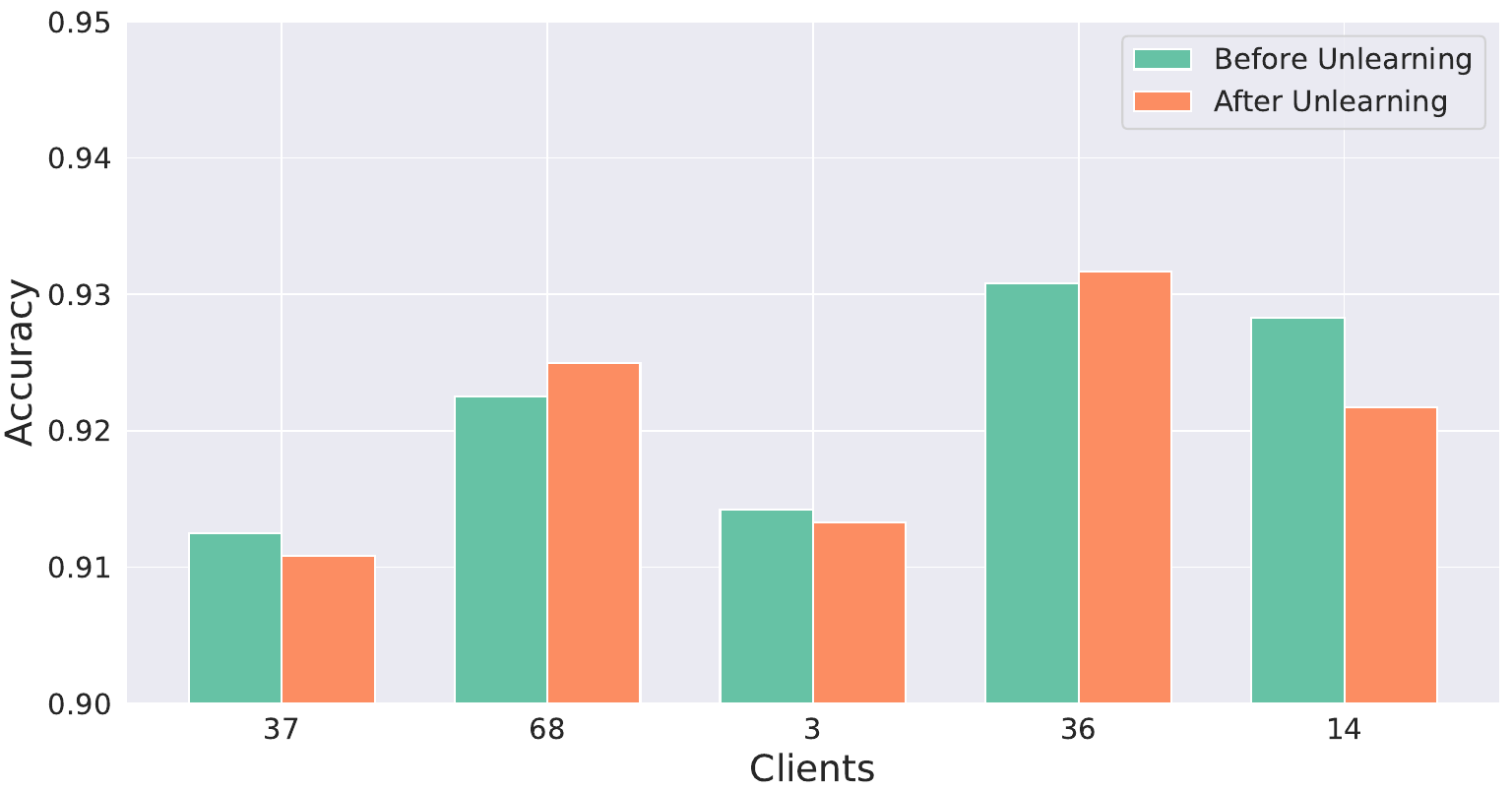}
        \caption{The comparison of 5 randomly chosen clients' performance before and after performing the unlearning procedure.}
        \label{fig:fu_clients}
    \end{subfigure}
    \caption{Performance comparisons of global model and individual clients before and after unlearning.}
    \label{fig:combined_figure}
\end{figure}

To validate our unlearning approach, we used the AG News dataset. We first trained FedDPG for 10 federated training rounds with a client selection ratio of 10\% and a prompt length of 10. We then randomly selected a participating client to simulate the unlearning process and followed the procedure described in Section \ref{sec:sub_unlearning}. For the unlearning experiment, we randomly mislabeled 20\% of the selected client's data. 
Figure \ref{fig:combined_figure} analyses the results of FedDPGu from two perspectives: global (Figure \ref{fig:fu_global}) and local (Figure \ref{fig:fu_clients}) performance. Specifically, Figure \ref{fig:fu_global} compares the accuracy of the global model at different stages, highlighting FedDPGu's minimal global performance degradation. Similarly, Figure \ref{fig:fu_clients} demonstrates that the newly unlearned model has only a marginal impact on the performance of other clients, even when evaluated on their private data.

\section{Conclusion}
In this paper, we introduced FedDPG, a flexible and dynamic PEFT technique for FL settings. Our experimental results demonstrated its superior performance compared to existing state-of-the-art methods across three text classification and sentiment analysis datasets. Additionally, we explored federated machine unlearning (FMU) and proposed FedDPGu, a promising approach to data unlearning that minimises the impact on global model performance in FL systems. While our initial results are encouraging, FedDPGu is in its early stages and requires further investigation with more robust unlearning verification methods, which we plan to explore in future work.
%
%
%
\bibliographystyle{splncs04}
\bibliography{references}

\end{document}